\documentclass{article}

\usepackage{microtype}
\usepackage{graphicx, caption, subcaption}
\usepackage{amsmath}
\usepackage{amssymb}
\usepackage{booktabs} 
\usepackage[square,numbers]{natbib}
\usepackage{multicol,lipsum,microtype}
\usepackage{dblfloatfix}
\usepackage{mathtools, nccmath}

\usepackage{hyperref}

\usepackage[accepted]{icml2021}

\begin{document}

\twocolumn[
\icmltitle{Unsupervised Skill-Discovery and Skill-Learning in Minecraft}



\icmlsetsymbol{equal}{*}

\begin{icmlauthorlist}
\icmlauthor{Juan José Nieto}{upc}
\icmlauthor{Roger Creus}{upc}
\icmlauthor{Xavier Giro-i-Nieto}{upc,bsc,iri}

\end{icmlauthorlist}

\icmlaffiliation{upc}{Universitat Polit\`{e}cnica de Catalunya, Barcelona, Spain}
\icmlaffiliation{bsc}{Barcelona Supercomputing Center, Barcelona, Spain}
\icmlaffiliation{iri}{Institut de Robòtica i Informàtica Industrial, CSIC-UPC, Barcelona, Spain}

\icmlcorrespondingauthor{Juan José Nieto}{juanjo.3ns@gmail.com}
\icmlcorrespondingauthor{Roger Creus}{creus99@protonmail.com}
\icmlcorrespondingauthor{Xavier Giro-i-Nieto}{xavier.giro@upc.edu}

\icmlkeywords{Reinforcement Learning, ICML, Unsupervised Learning, Skill-discovery, self-supervised learning, intrinsic motivation, empowerment}

\vskip 0.3in
]



\printAffiliationsAndNotice{}  

\begin{abstract}
Pre-training Reinforcement Learning agents in a task-agnostic manner has shown promising results. However, previous works still struggle in learning and discovering meaningful skills in high-dimensional state-spaces, such as pixel-spaces. We approach the problem by leveraging unsupervised skill discovery and self-supervised learning of state representations. In our work, we learn a compact latent representation by making use of variational and contrastive techniques. We demonstrate that both enable RL agents to learn a set of basic navigation skills by maximizing an information theoretic objective. We assess our method in Minecraft 3D pixel maps with different complexities. Our results show that representations and conditioned policies learned from pixels are enough for toy examples, but do not scale to realistic and complex maps. To overcome these limitations, we explore alternative input observations such as the relative position of the agent along with the raw pixels.
\end{abstract}


\section{Introduction}

Reinforcement Learning (RL)~\cite{sutton2018reinforcement} has witnessed a wave of outstanding works in the last decade, with special focus on games (\citet{schrittwieser2020mastering}, \citet{vinyals2019grandmaster}, \citet{berner2019dota}), but also in robotics (\citet{akkaya2019solving}, \citet{hwangbo2017control}). In general, these works follow the classic RL paradigm where an agent interacts with an environment performing some action, and in response it receives a reward. These agents are optimized to maximize the expected sum of future rewards. 

Rewards are usually handcrafted or overparametrized, and this fact becomes a bottleneck that prevents RL to scale. For this reason, there has been an increasing interest in training agents in a task-agnostic manner during the last few years, making use of intrinsic motivations and unsupervised techniques. 
Recent works have explored the unsupervised learning paradigm (\citet{campos2020explore, gregor2016vic, eysenbach2018diayn,warde2018discern, burda2018rnd, pathak2017icm}), but RL is still far from the remarkable results obtained in other domains. For instance, in computer vision, \citet{chen2021mocov3} achieve an 81\% accuracy on ImageNet training in a self-supervised manner, or \citet{caron2021dino} achieves state-of-the-art results in image and video object segmentation using Visual Transformers \cite{dosovitskiy2020vit} and no labels at all. Also, in natural language processing pre-trained language models such as GPT-3 \cite{brown2020gpt3} have become the basis for other downstream tasks. 

Humans and animals are sometimes guided through the process of learning. We have good priors that allow us to properly explore our surroundings, which leads to discovering new skills. For machines, learning skills in a task-agnostic manner has proved to be challenging~\cite{warde2018discern,lee2019slac}. These works state that training pixel-based RL agents end-to-end is not efficient because 
learning a good state representation is unfeasible due to the high dimensionality of the observations. Moreover, most of the successes in RL come from training agents during thousands of simulated years (\citet{berner2019dota}) or millions of games (\citet{vinyals2019grandmaster}). 
This learning approach is very sample inefficient and 
sometimes limits its research because of the high computational budget it may imply.
As a response, some benchmarks have been proposed to promote the development of algorithms that can reduce the number of samples needed to solve complex tasks. This is the case of 
MineRL (\citet{guss2021minerl}) or ProcGen Benchmark (\citet{cobbe2019procgen}).

Our work is inspired by \citet{campos2020explore} and their \textit{Explore, Discover and Learn}~(EDL) paradigm. EDL relies on \textit{empowerment} \cite{salge2014empowerment} for motivating an agent intrinsically. Empowerment aims to maximize the influence of the agent over the environment while discovering novel skills. As stated by \citet{salge2014empowerment}, this can be achieved my maximizing the mutual information between sequences of actions and final states. \citet{gregor2016vic} introduces a novel approach that instead of blindly committing to a sequence of actions, each action depends on the observation from the environment. This is achieved by maximizing the mutual information between inputs and some latent variables. \citet{campos2020explore} embraces this approach as we do. However, the implementation by \citet{campos2020explore} makes some assumptions that are not realistic for pixel observations. Due to the Gaussian assumption at the output of variational approaches, the intrinsic reward is computed as the reconstruction error, and in the pixel domain this metric does not necessarily match the distance in the environment space. Therefore, we look for alternatives that suit our requirements: we derive a different reward from the mutual information, and we study alternatives to the variational approach.

This work focuses on learning meaningful representations, discovering skills and training latent-conditioned policies. In any of the cases, our methodology does not require any supervision and works directly from pixel observations. Additionally, we also study the impact of extra input information in the form of position coordinates. Our proposal is tested over the MineRL~\cite{guss2021minerl} environment, which is based on the popular Minecraft videogame. Even though the game proposes a final goal, Minecraft is well known by the freedom that it gives to the players, and actually most human players use this freedom to explore this virtual world following their intrinsic motivations.
Similarly, we aim at discovering skills in Minecraft without any extrinsic reward. 

We generate random trajectories in Minecraft maps with little exploratory challenges, and also study contrastive alternatives that exploit the temporal information throughout a trajectory. 
The contrastive approach aims at learning an embedding space where observations that are close in time are also close in the embedding space. A similar result can be achieved by leveraging the agents' relative position in the form of coordinates. In the latter, the objective is to infer skills that do not fully rely on pixel resemblance, but also take into account temporal and spatial relationships.

Our final goal is to discover and learn skills that can be potentially used in more broad and complex tasks. Either by transferring the policy knowledge or by using hierarchical approaches. Some works have already assessed this idea specially in robotics~\cite{florensa2017stochastic} or 2D games~\cite{campos2021coverage}. Once the pre-training stage is completed and the agent has learned some basic behaviours or skills, the agent is exposed to an extrinsic reward. These works show how the agents leverage the skill knowledge to learn much faster and encourage proper exploration of the environment in unrelated downstream tasks. However, transferring the policy knowledge is not as straightforward as in other deep learning tasks. If one wants to transfer behaviours (policies), the change in the task might lead to catastrophic forgetting.

Our contributions are the following:
\begin{itemize}
    \item We demonstrate that variational techniques are not the only ones capable of maximizing the mutual information between inputs and latent variables by leveraging contrastive techniques.
    \item We provide alternatives for discovering and learning skills in procedurally generated maps by leveraging the agents coordinate information. 
    \item We succesfully implement the reverse form of the mutual information for optimizing pixel-based agents in a complex 3D environment.
\end{itemize}

\section{Related Work}
\label{sec:sota}

Intrinsic Motivations (IM) are very helpful mechanisms to deal with sparse rewards. In some environments the extrinsic rewards are very difficult to obtain and, therefore, the agent does not receive any feedback to progress. In order to drive the learning process without supervision, we can derive intrinsic motivations as proxy rewards that guide the agents towards the extrinsic reward or just towards better exploration.

\textbf{Skill Discovery.} We relate Intrinsic Motivations to the concept of \textit{empowerment}~\cite{salge2014empowerment}, a RL paradigm in which the agent looks for the states where it has more control over the environment. \citet{mohamed2015variational} derived a variational lower bound on the mutual information that allows to maximize empowerment. Skill discovery extends this idea from the action level to temporally-extended actions.
\citet{florensa2017stochastic} merges skill discovery and hierarchical architectures. They learn a high-level policy on top of some basic skills learned in a task-agnostic way. They show how this set-up improves the exploration and enables faster training in downstream tasks. Similarly, \citet{achiam2018variational} emphasize in learning the skills dynamically using a curriculum learning approach, allowing the method to learn up to a hundred of skills. Instead of maximizing the mutual information between states and skills they use skills and whole trajectories. \citet{eysenbach2018diayn} demonstrates that learned skills can serve as an effective pretraining mechanism for robotics. Our work follows their approach regarding the use of a categorical and uniform prior over the latent variables. \citet{campos2020explore} exposes the lack of coverage of previous works. They propose \textit{Explore, Discover and Learn} (EDL), a method for skill discovery that breaks the dependency on the distributions induced by the policy. \citet{warde2018discern} provides an algorithm for learning goal-conditioned policies using an imitator and a teacher. They demonstrate the effectiveness of their approach in pixel-based environments like Atari, DeepMind Control Suite and DeepMind Lab.

\textbf{Intrinsic curiosity.} In a more broad spectrum we find methods that leverage intrinsic rewards that encourage exploratory behaviours. \citet{pathak2017icm} present an \textit{Intrinsic Curiosity Module} that defines the curiosity or reward as the error predicting the consequence of its own actions in a visual feature space. Similarly, \citet{burda2018rnd} uses a Siamese network where one of the encoder tries to predict the output of the other. The bonus reward is computed as the error between the prediction and the random one. 

\textbf{Goal-oriented RL.} Many of the works dealing with skill-discovery end up parameterizing a policy. This policy is usually conditioned in some goal or latent variable $z\sim Z$. The goal-conditioned policy formulation along with function approximation was introduced by \citet{schaul2015universal}
\textit{Hindsight Experience Replay} (HER) by \citet{andrychowicz2017her}
 allows sample-efficient learning from environments with sparse rewards. HER assumes that any state can be a goal state. Therefore, leveraging this idea, the agent learns from failed trajectories as if the final state achieved was the goal state. \citet{trott2019keeping} proposes to use pairs of trajectories that encourage progress towards a goal while learns to avoid local optima. Among different environments, \textit{Sibling Rivalry} is evaluated in a 3D construction task in Minecraft. The goal of the agent is to build a structure by placing and removing blocks. They use the number of block-wise differences as a naive shaped reward which causes the agent to avoid placing blocks. Simply by adding Sibling Rivalry they manage to improve the degree of construction accuracy.

\textbf{Representation Learning in RL.} In RL we seek for low-dimensional state representations that preserve all the information and variability of the state space in order to make decisions that eventually maximize the reward. This becomes crucial when dealing with pixel-based environments.
 \citet{ghosh2018learning} aims to capture those factors of variation that are important for decision making and are aware of the dynamics of the environment without the need for explicit reconstruction. \citet{lee2019slac} also makes use of variational inference techniques for estimating the posterior distribution, but breaks the Markovian assumption by conditioning the probability of the latent variables with past observations. \citet{oord2018cpc} leverages powerful autoregressive models and negative sampling to learn a latent space from high-dimensional data. This work introduces the InfoNCE loss based on Noise Contrastive Estimation. The intuition behind is that we learn a latent space that allows to classify correctly positive pairs while discriminating from negative samples. We make use of this loss in our contrastive experiments, as well as the following works explained. \citet{srinivas2020curl} trains an end-to-end model by performing off-policy learning on top of extracted features. This features are computed using a contrastive approach based on pairs of augmented observations, while \citet{stooke2020atc} picks pairs of delayed observations.
 All these works learn representations from pixel-based observations, except for the first one by \citet{ghosh2018learning}.

\section{Information-theoretic skill discovery}
\label{sec:background}

Intrinsic Motivations (IM) can drive the agents learning process in the absence of extrinsic rewards. With IM, the agents do not receive any feedback from the environment, but must autonomously learn the possibilities available in it. To achieve that, the agents aim to gain resources and influence on what can be done in the environment. In the \textit{empowerment} \cite{salge2014empowerment} framework, an agent will look for the state in which has the most control over the environment. This concept usually deals with simple actions. In contrast, \textit{skill-discovery} temporally abstracts these simple actions to create high-level actions that are dubbed as \textit{skills} or \textit{options}. Skill-discovery is formulated as maximizing the mutual information between inputs and skills. This encourages the agent to learn skills that derive as many different input sequences as possible, while avoiding overlap between sequences guided by different skills. Therefore, \textit{skill-discovery} can ease the exploration in complex downstream tasks. Instead of executing random actions, the agent can take advantage of the learned behaviours in order to perform smarter moves towards states with potential extrinsic rewards.

In the next section we formulate the mathematical framework that is used through our work. First we define the classic Markov decision process that typically provide a mathematical formulation for reinforcement learning tasks. Later on, we introduce the tools from information-theory that allows us to define the maximization of the mutual information.

\subsection{Preliminaries}
\label{subsec:preliminaries}

Let us consider a Markov decision process (MDP) without rewards as $\mathcal{M} = (\mathcal{S}, \mathcal{A}, \mathcal{P})$. Where $\mathcal{S}$ is the high-dimensional state space (pixel images),  $\mathcal{A}$ refers to the set of actions available in the environment and $\mathcal{P}$ defines the transition probability $p(s_{t+1}|s_t,a)$. We learn latent-conditioned policies $\pi(a|s,z)$, where the latent $z \in \mathcal{Z}$ is a random variable. 

Given the property of symmetry, the mutual information ($\mathcal{I}$) can be defined using the Shannon Entropy ($\mathcal{H}$) in two ways (\citet{gregor2016vic}):

\begin{equation}
    \begin{aligned}
        \mathcal{I}(S,Z) & = \mathcal{H}(Z) -\mathcal{H}(Z|S) \qquad \rightarrow \qquad \text{reverse} \\
                        & = \mathcal{H}(S) -\mathcal{H}(S|Z) \qquad \rightarrow \qquad \text{forward}
    \end{aligned}
\end{equation}

For instance, we derive the reverse form of the mutual information (MI):

\begin{equation}
    \mathcal{I}(S,Z) = \mathbb{E}_{s,z \sim p(z,s)}[\log p(z|s)] - \mathbb{E}_{z\sim p(z)}[\log p(z)]
    \label{eq:reverseMI}
\end{equation}

The posterior $p(z|s)$ is unknown due to the complexity to marginalize the evidence $p(s) = \int p(s|z)p(z)dz$. Hence, we approximate it with $q_\phi(z|s)$ by using variational inference or contrastive techniques. As a result, we aim at maximizing a lower bound based on the KL divergence. For a detailed derivation we refer the reader to the original work from \citet{mohamed2015variational}.

Therefore, the MI lower bound becomes:

\begin{equation}
    \mathcal{I}(S,Z) \ge \mathbb{E}_{s,z \sim p(z,s)}[\log q_\phi(z|s)] - \mathbb{E}_{z\sim p(z)}[\log p(z)]
    \label{eq:mutual_information}
\end{equation}

The prior $p(z)$, can either be learned through the optimization process \cite{gregor2016vic}, or fixed uniformly beforehand \cite{eysenbach2018diayn}. In our case, since we want to maximize the uncertainty over $p(z)$, we define a categorical uniform distribution.

\subsubsection{Variational Inference}

Variational inference is a well-known method for modelling posterior distributions. Its main advantage is that it avoids computing the marginal $p(s)$, which is usually impossible. Instead, one selects some tractable families of distributions $q$ as an approximation of $p$. 

\begin{equation}
    p(z|s) \approx q(z|\theta)
\end{equation}

We fit q with sample data to learn the distribution parameters $\theta$.
In particular, we make use of Variational Auto-Encoders (VAE) (\citet{kingma2014vae}) with categorical classes, namely VQ-VAE \cite{oord2017vqvae}, for modelling the mappings from inputs to latents and backwards. The encoder $q_\phi(z|s)$ models $p(z|s)$ and the decoder $q_\psi(s|z)$ models $p(s|z)$.

\subsubsection{Contrastive Learning}

Contrastive learning is a subtype of self-supervised learning that consists in learning representations from the data by comparison.  This field has quickly evolved in the recent years, especially in computer vision. Within this context, the comparisons are between pairs of images. Examples of positive pairs may be augmented versions of a given image like crops, rotations, color transformations, etc; and the negative pairs maye be the other images from the dataset.
In this case, representations are learned by training deep neural networks to distinguish between positive or negative pairs of observations.
The main diference between contrastive self-supervised (or unsupervised) learning and metric learning is that the former does not require any human annotation, while the later does.
Learning representations with no need of labeling data allows scaling up the process and overcoming the main bottleneck when training deep neural networks: the scarcity of supervisory signals.


Similar to what is done in variational techniques, we can compute the mutual information between inputs and latents. In this case, instead of learning the latents by the reconstruction error, they are learned by modeling global features between positive and negative pairs of images. Intuitively, two distinct augmentations (positive pair) should be closer in the embedding space than two distinct images (negative pair) from the dataset. Therefore, we need two encoders in parallel instead of an encoder-decoder architecture. The second encoder is usually called momentum encoder (\citet{chen2021mocov3}) and its weights are updated using exponential moving averages of the main encoder.

In each training step, we forward a batch of different original images through the main encoder while we forward the positive pairs of each of those images through the \textit{momentum} encoder. In a batch of $N$ samples, we have for each positive pair, $N-1$ negative pairs. We define $z$ as the output of the encoder and $z'$ as the output of the momentum encoder. Then, $e$ is a convolutional neural encoder and $h$ is a projection head (small multi-layer perceptron) that returns a latent $z$, ($z = h(e_\theta(s))$).


At the output of the network we can perform a categorical cross-entropy loss where the correct classes are the positive pairs in the batch. The correct classes are in the diagonal of the resulting matrix $z^TWz'$, where $W$ is a projection matrix learned during training. In the other positions, we have the similarity between negative pairs in the embedding space.
Minimizing this loss, whose name is InfoNCE \cite{oord2018cpc}, will encourage the model to find global features that can be found in augmented versions from an image. Moreover, as stated by their authors \citet{oord2018cpc}, minimizing the InfoNCE Eq.~\ref{eq:infoNCE} loss maximizes the mutual information between inputs and latents. As a result, we learn the desired mapping from input states $s$ to latent vectors $z$.

\begin{equation}
    \mathcal{L}_{InfoNCE} = \log \frac{\exp(z^TWz')}{\exp(z^TWz') + \sum_{i=0}^{K-1} \exp(z^TWz_i')}
    \label{eq:infoNCE}
\end{equation}



\section{Methodology}
\label{sec:methodology}

In this section we show the implementation details adopted for each of the stages of our method.

\subsection{Exploration}
\label{sec:exploration}

EDL (Explore, Discover and Learn)~\cite{campos2020explore} provides empirical and theoretical analysis of the lack of coverage of some methods leveraging the mutual information for discovering skills. Either by using the forward or the reverse form of the MI, the reward given to novel states is always smaller than the one given to known states. The main problem dwells in the induced state distributions used to maximize the mutual information. Most works reinforce already discovered behaviours since they induce the state distribution from a random policy, $p(s) \approx \rho_\pi(s) = \mathbb{E}_z[\rho_\pi(s|z)]$. EDL is agnostic to how $p(s)$ is obtained, so we can infer it in very different ways. One manner would be to induce it  by leveraging a dataset of expert trajectories that will encode human-priors that are usually learned poorly with information-theoretic objectives. Another way is to make use of an exploratory policy that is able to induce a uniform distribution over the state space. Since this is not clearly solved in complex 3D environments yet, we explore by using random policies in bounded maps.

\subsection{Skill-Discovery}

\begin{figure}[!hbt]
    \centering
    \includegraphics[width=\columnwidth]{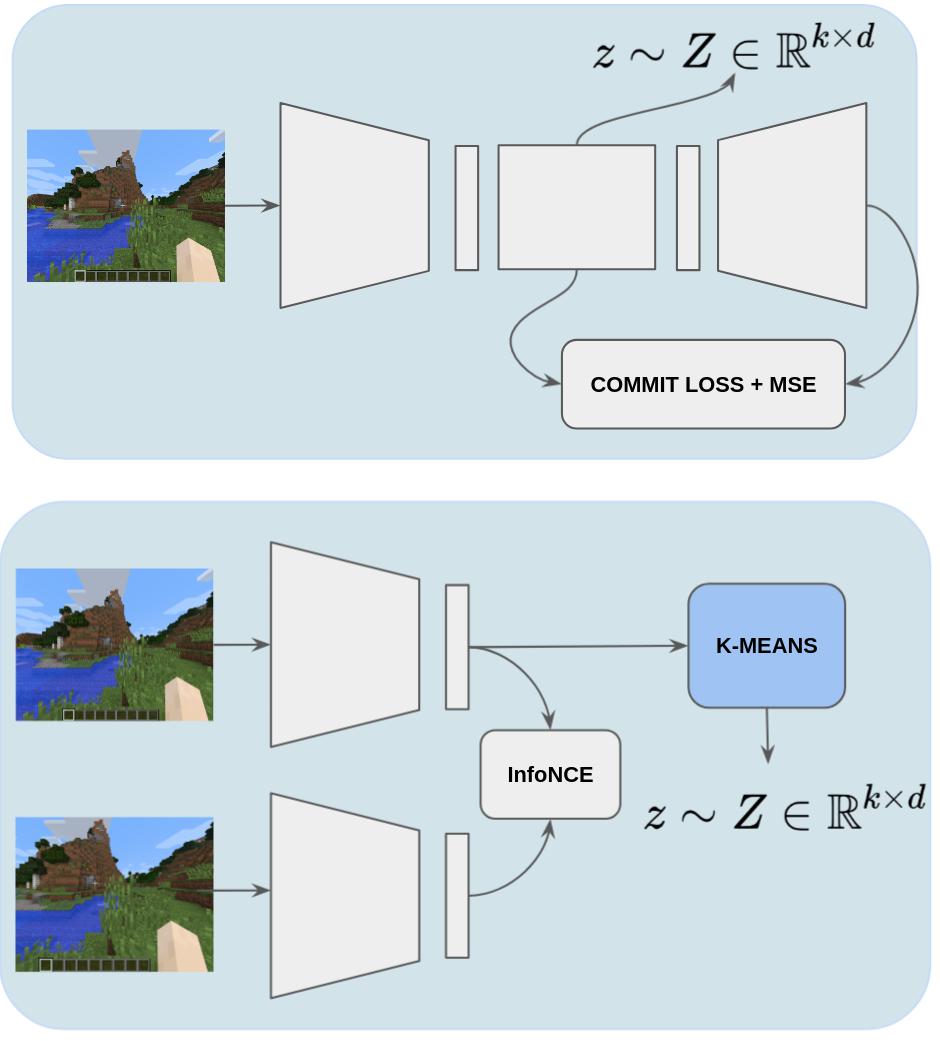}
    \caption{Skill-discovery pipeline with (top) variational and (bottom) contrastive approaches.}
    \label{fig:skill-discovery}
\end{figure}

In Section~\ref{sec:background}, we proposed two distinct approaches for modelling the mapping from the observations $s$ to the latent variables $z$: variational inference and contrastive learning.
The skill-discovery pipelines for both approaches are depicted in Figure \ref{fig:skill-discovery}, and are described in the remaining of this section.


\subsubsection{Variational Inference}
\label{sec:variational-discovery}

Vector-Quantized VAE (VQ-VAE)~\cite{oord2017vqvae} is a variational model that allows us to have a categorical distribution $p(z)$ as a bottleneck (also called codebook), an encoder $q_\psi$ that estimates the posterior $p(z|s)$, and the decoder $q_\phi$ that estimates $p(s|z)$.

Before training, the model requires to fix the length of the codebook. This determines the granularity of the latent variables. If we choose a large number of codes, we will end up with latent variables that encode very similar states. Instead, if we choose a small number, we will encourage the model to find latents that generalize across diverse scenarios. The perplexity metric measures the number of codewords needed to encode our whole state distribution. In practice, this metric is computed per batch during training. Since it is not possible to know beforehand the number of useful latent variables that our model will discover, one can iterate over different codebook lengths until they find a good trade-off between generalization and granularity.

In EDL~\cite{campos2020explore}, the purpose of training a variational auto-encoder was to discover the latent variables that condition the policies in the Learning stage. But the learned representations were discarded. Instead, in our case, we also leverage the representations learned in the encoder and decoder for training the RL agent. This allows for faster and more efficient trainings of the RL agents.

\subsubsection{Contrastive Learning}
\label{sec:contrastive-discovery}

In the contrastive case, we follow the idea of \citet{stooke2020atc} where the positive pairs are delayed observations from a specific trajectory. Once the latents are learned, we define a categorical distribution over latents by clustering the embedding space of the image representations \cite{yarats2021reinforcement}. This step is performed using K-Means with $K$ clusters equal to the length of the VQ-VAE codebook.

At this point, we have the same set-up for both the contrastive and variational approaches. The K-Means centroids are equivalent to the VQ-VAE codebook embeddings, so we can maximize the mutual information between the categorical latents and the inputs with the same Equation~\ref{eq:reverseMI}.

\subsection{Skill-Learning}

\begin{figure}[!hbt]
    \centering
    \includegraphics[width=0.85\columnwidth]{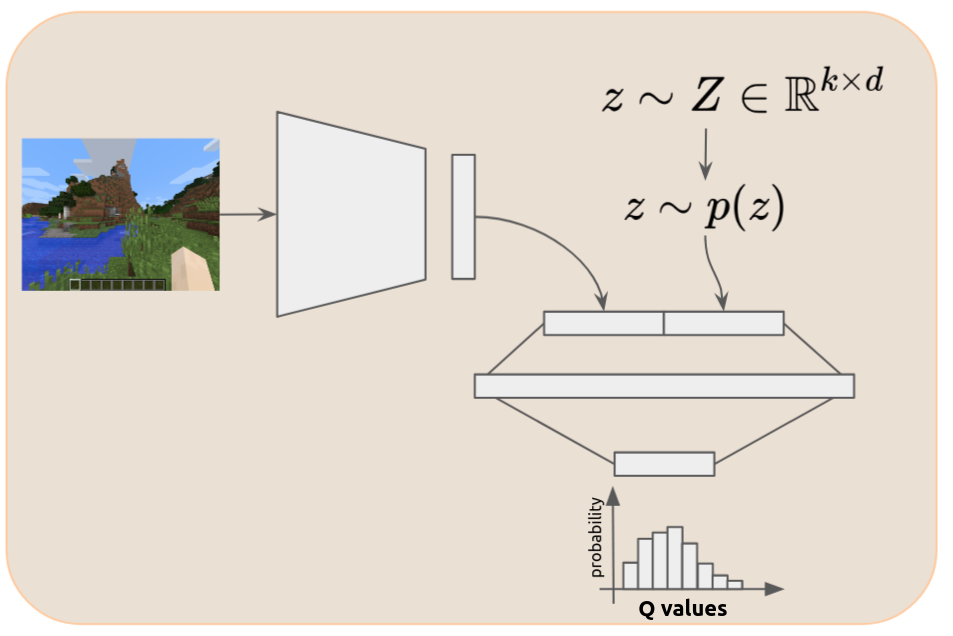}
    \caption{Skill-Learning pipeline for both contrastive and variational approaches.}
    \label{fig:skill-learning}
\end{figure}

In the last stage of the process, we aim to train a policy $\pi(a|s,z)$ that maximizes the mutual information between inputs and the discovered latent variables. At each episode of the training, we sample a latent variable $z\sim p(z)$. Then, at each step, this embedding is concatenated with the embedding of the encoded observation at timestep $t$ and forwarded through the network. This process is depicted in Figure~\ref{fig:skill-learning}. At this stage, the latent variables are interpreted as navigation goals. Hence, the agent is encouraged to visit those states that brings it closer to the navigation goal or latent variable. Since each of the latent variables encode different parts of the state distribution, this results in covering different regions for each $z$.
Our method adopts the reverse form of the mutual information. Since we fix $p(z)$ as a uniform distribution we can remove the $\log p(z)$ constant term from the Equation~\ref{eq:mutual_information}, which results in the reward of Equation~\ref{eq:reward_reverseMI_v} for the variational case, and Equation~\ref{eq:reward_reverseMI_c} for the contrastive case. 
The agent receives a reward of 1 only if the embedding that is conditioning the policy is the closest to the encoded observation.
\begin{equation}
    r(s,z) = q_\phi(z|s)
    \label{eq:reward}
\end{equation} 

\begin{equation}
    q_\phi(z=k|s)=
        \begin{cases}
            1, &  \text{if} \quad k=argmin_j ||z_e(s)-e_j||\\
            0,              & \text{otherwise}
        \end{cases}
    \label{eq:reward_reverseMI_v}
\end{equation}

Where $z_e(s)$ is the encoded observation and $e$ is the codebook of embeddings.

\begin{equation}
    q_\phi(z=k|s)= 
        \begin{cases}
            1, & \text{for } k=argmax_j (z_e(s)We_j)\\
            0,              & \text{otherwise}
        \end{cases}
    \label{eq:reward_reverseMI_c}
\end{equation}

In the contrastive case, we compute a similarity measure. Therefore, we look for the latent index with highest similarity to the current observation.

\section{Experiments}
\label{sec:results}

In this section we asses our method in two different Minecraft maps and set-ups. In each of the trainings we follow the pipeline described in Section~\ref{sec:methodology}, where we first generate random trajectories in the specified map. Later, we discover some categorical latents from them, and finally we train some conditioned policies on these latents. We refer the reader to the Appendix~\ref{sec:hyperparams} to check the hyperparameters used in the different methods.

\subsection{Discovery and learning of skills in handcrafted maps}
\label{ssec:map-handcrafted}

\begin{figure}[!hbt]
    \centering
    \begin{subfigure}[!hbt]{0.4\columnwidth}
        \includegraphics[width=\textwidth]{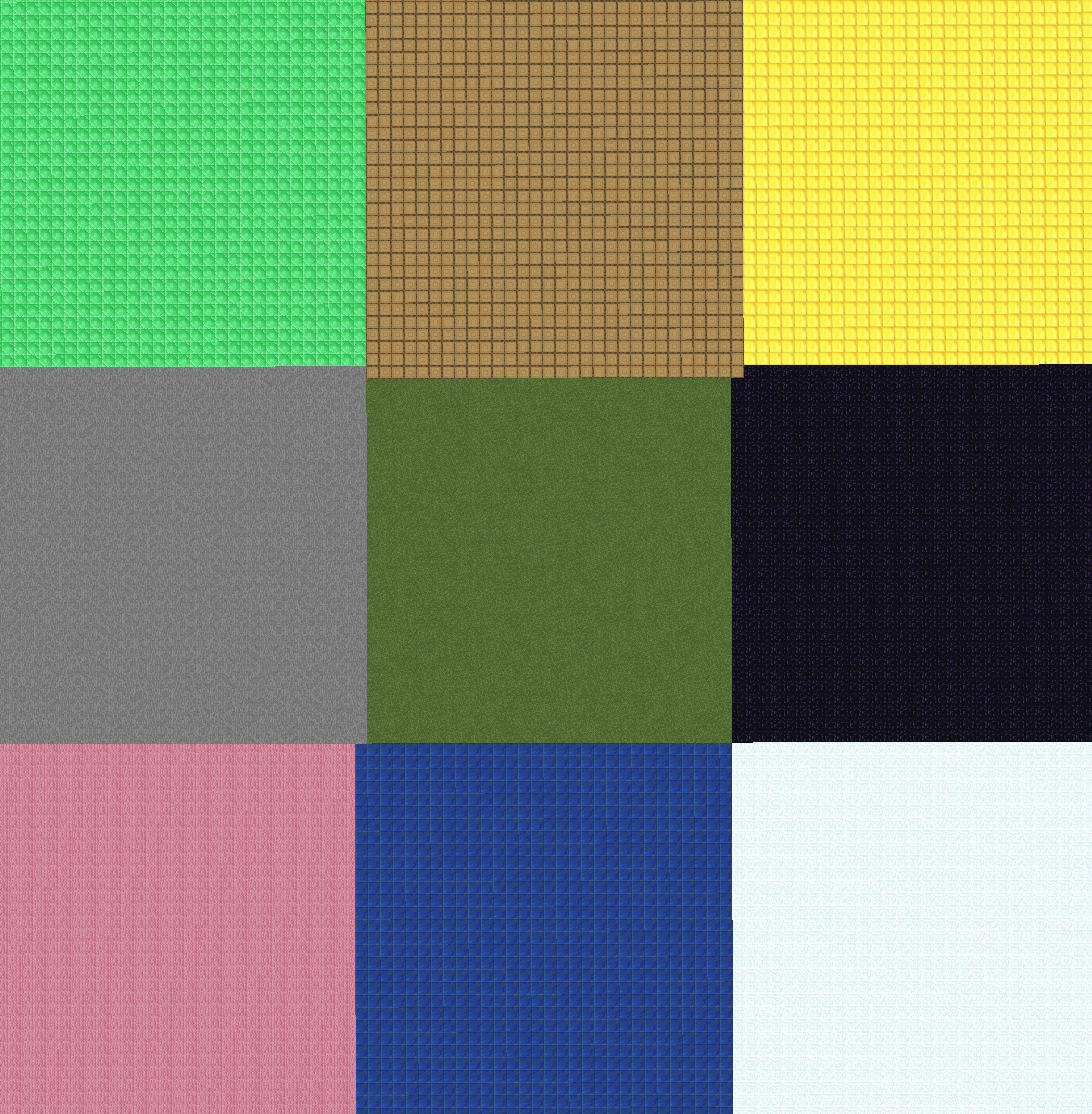}    
        \caption{}
    \end{subfigure}
    \begin{subfigure}[!hbt]{0.45\columnwidth}
        \includegraphics[width=\textwidth]{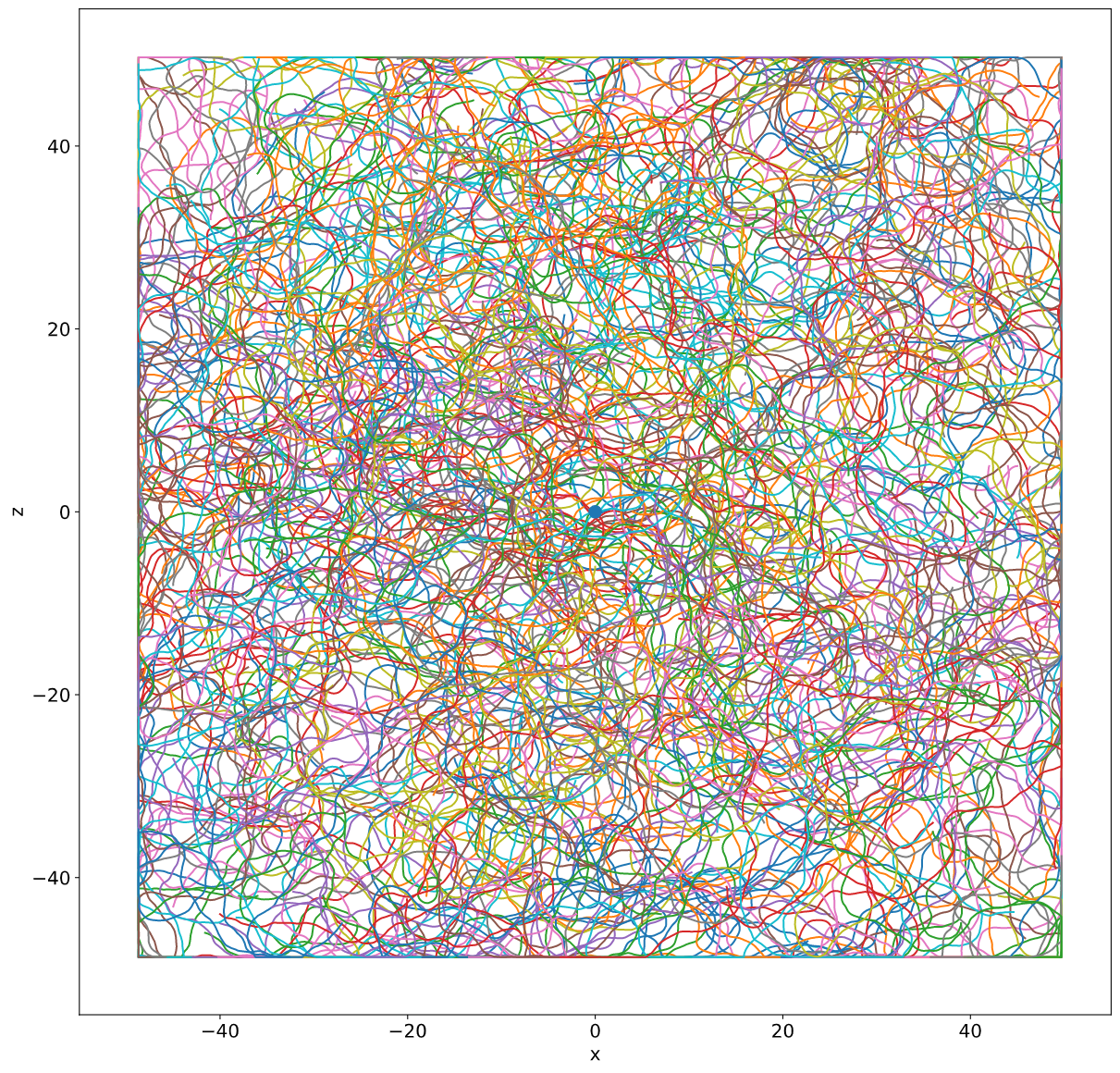}    
        \caption{}
    \end{subfigure}
    \caption{\textbf{(a)} Top-view of the toy map used for asses our method. The dimensions of this map are 100 square meters, if we consider that the Minecraft coordinate system was measured in meters. The agent has no exploratory issues even with a random policy. \textbf{(b)} Trajectories generated by a random agent deployed in any region of the map. }
    \label{fig:toymap}
\end{figure}

The first map, in Figure~\ref{fig:toymap}a, is a 2D handcrafted map that consists in 9 different regions where the only difference among them is the floor type. The $p(s)$ induced from the random trajectories seen in Figure~\ref{fig:toymap}b covers the state subspace uniformly.

\begin{figure}[!b]
    \centering
    \includegraphics[width=\columnwidth]{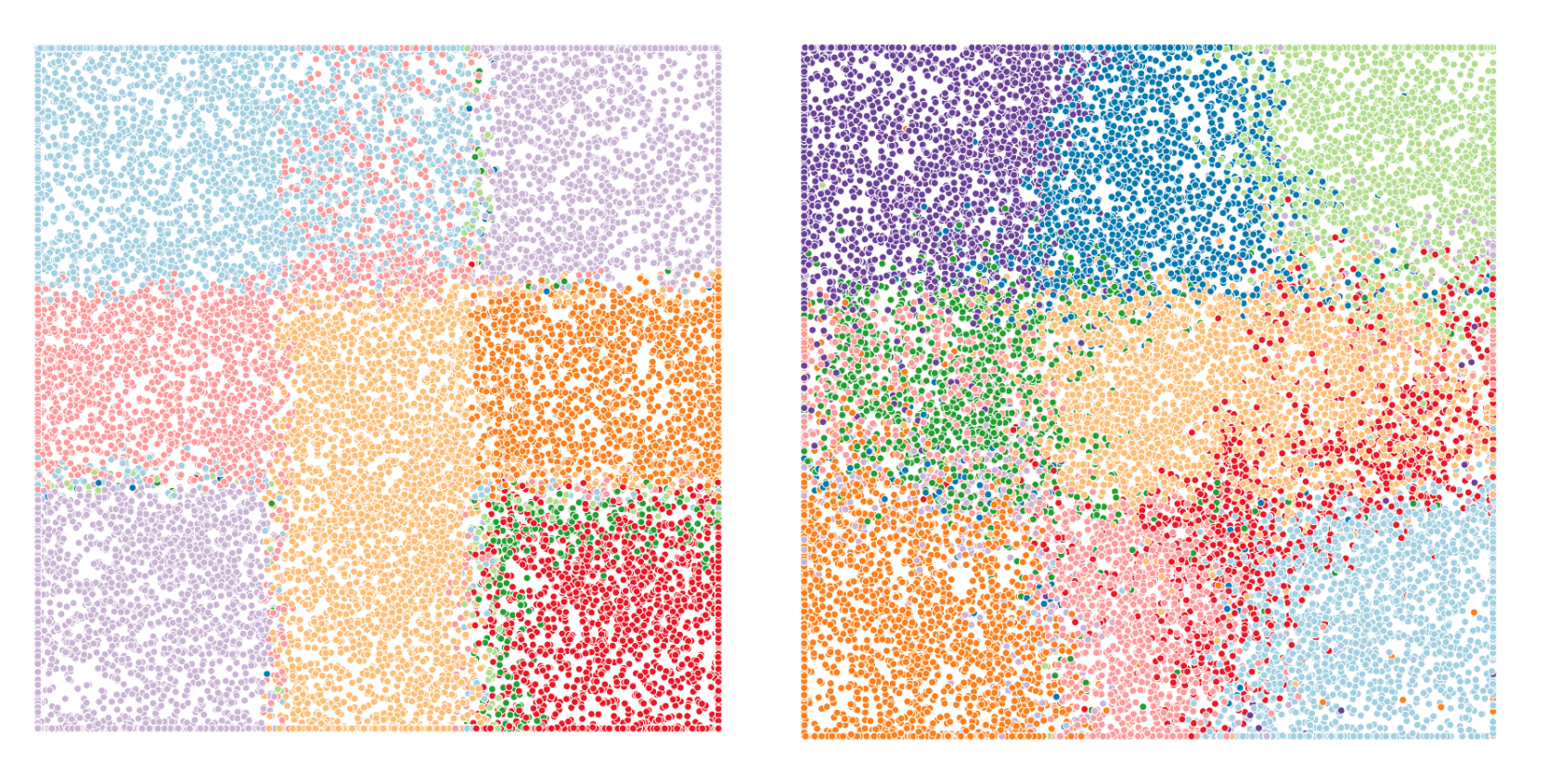}
    \caption{\textbf{Left:} Index map generated by the variational approach. \textbf{Right:} Index map generated by the contrastive approach.}
    \label{fig:toy_indexmaps}
\end{figure}

\begin{figure*}[!hbt]
    \centering
    \begin{subfigure}[!hbt]{0.28\textwidth}
        \includegraphics[width=\textwidth]{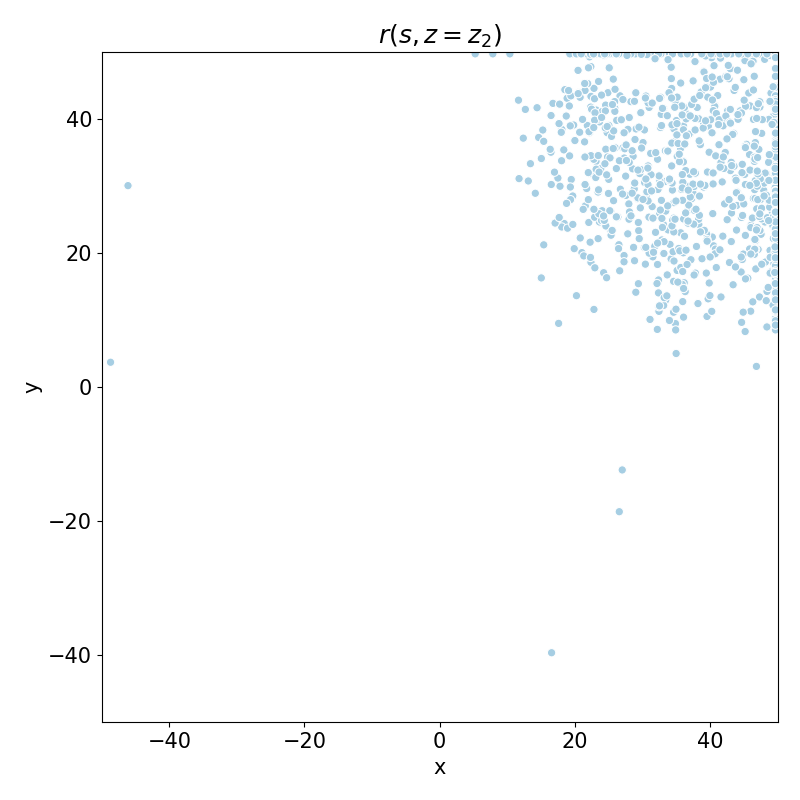}    
        \caption{}
    \end{subfigure}
    \begin{subfigure}[!hbt]{0.28\textwidth}
        \includegraphics[width=\textwidth]{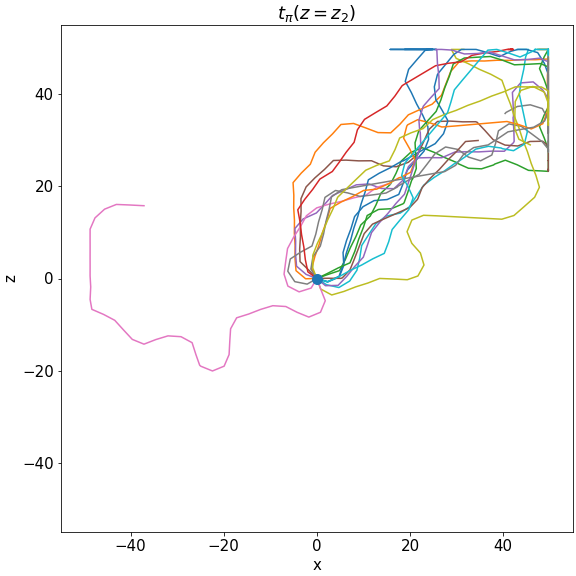}    
        \caption{}
    \end{subfigure}
    \begin{subfigure}[!hbt]{0.28\textwidth}
        \includegraphics[width=\textwidth]{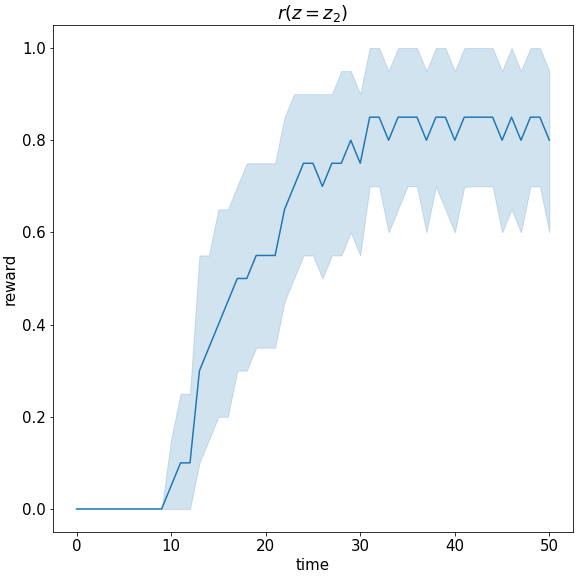}    
        \caption{}
    \end{subfigure}

    \caption{\textbf{(a)} Observations that are encoded as the latent vector 2. \textbf{(b)} shows the trajectories performed by the agent and \textbf{(c)} the average reward received in those trajectories. Both plots conditioning the agent on the latent vector 2. }
    \label{fig:skill2}
\end{figure*}

The \textit{index maps} shown in Figure~\ref{fig:toy_indexmaps} are generated from the random trajectories in Figure~\ref{fig:toymap}b. We encode each observation from each trajectory with the index of the closest embedding in the latent codebook or K-Means centroids. We assign a color to each of the indexes and we get what we refer as an \textit{index map}. Thanks to these maps, it is very easy to visualize whether the learned embeddings are meaningful or not.

Figure~\ref{fig:toy_indexmaps} compares the index maps learned with the variational and constrastive methods. In general we can see that the variational model discretizes slightly better, however it is unable to distinguish between two particular regions. Instead, despite the mixing and overlapping among different discovered latents, the contrastive approach manages to clusterize these regions decently. When studying these visualizations we should be aware that we are not taking into account the direction in which the agent was looking at the time it was recorded. Therefore, two points that are next to each other, might be encoding totally different views and hence will result in different latent codes.

Learning representations from these observations is quite straightforward, since the dynamics are not very complex. Despite that, either when training with the contrastive and variational methods, we usually find latent centroids that encode observations in between plots. For this reason, if we wanted to have a categorical latent vector for each of the regions we could overclusterize the latent embeddings (in the contrastive case) or select a larger codebook (in the variational case). Since we train the agent in a map where we have already collected observations from random policies, we can plot the reward that those observations would be given when conditioned in each of the categorical latents. For instance, in Figure~\ref{fig:skill2}a we can see that the observations recorded in the top-right corner are encoded as the second discovered latent. Therefore, if we condition the policy with it, the agent will receive a reward of 1, following Equation~\ref{eq:reward}.

In Figure~\ref{fig:skill2}b, we show how the performance of the agent when conditioned in this skill. All the trajectories head to the navigation goal except for one. Moreover, regardless of the initial direction, we can see that the agent is capable of orientating and go to the right direction. Lastly, in Figure~\ref{fig:skill2}c, we can see how it requires 10 steps to start triggering the intrinsic reward until reaching a stable 0.8 reward in average. For more examples on other latent codes we refer the reader to the Appendix~\ref{sec:full_toyexample}.

\begin{figure*}[!hbt]
    \centering
    \includegraphics[width=.9\textwidth]{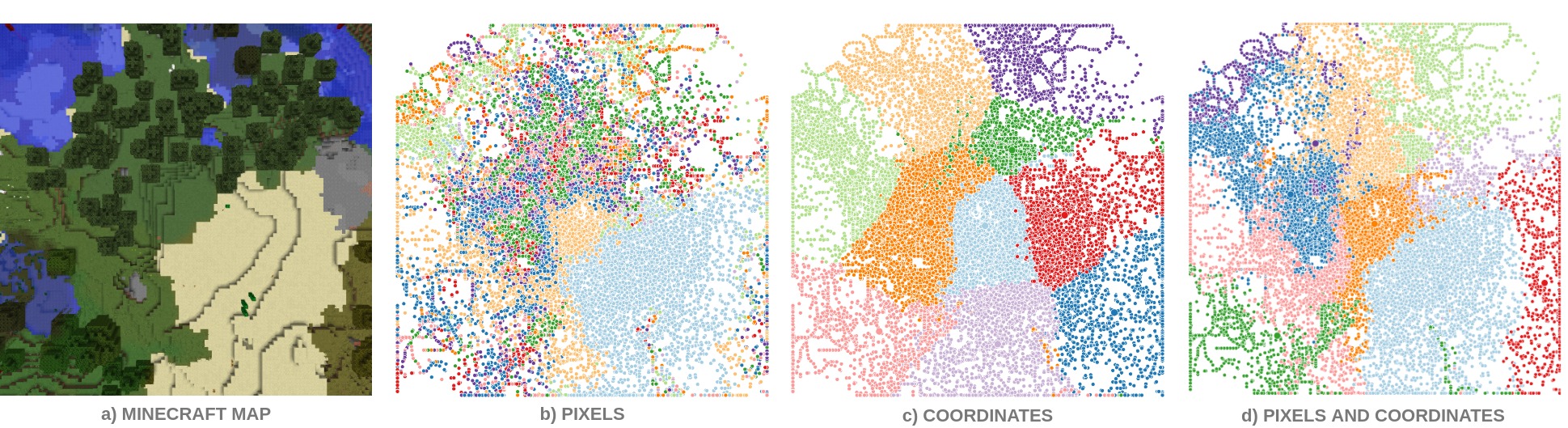}
    \caption{a) Top-view of our Minecraft map. The caption below b) c) d) refers to the nature of the states (pixels, coordinates or both) where all the trajectories were collected by random policies.
    Each coloured point indicates the closest centroid to the encoded embedding.}
    \label{fig:indexmaps}
\end{figure*}

\begin{figure*}[!b]
    \centering
    \begin{subfigure}[!hbt]{0.28\textwidth}
        \includegraphics[width=\textwidth]{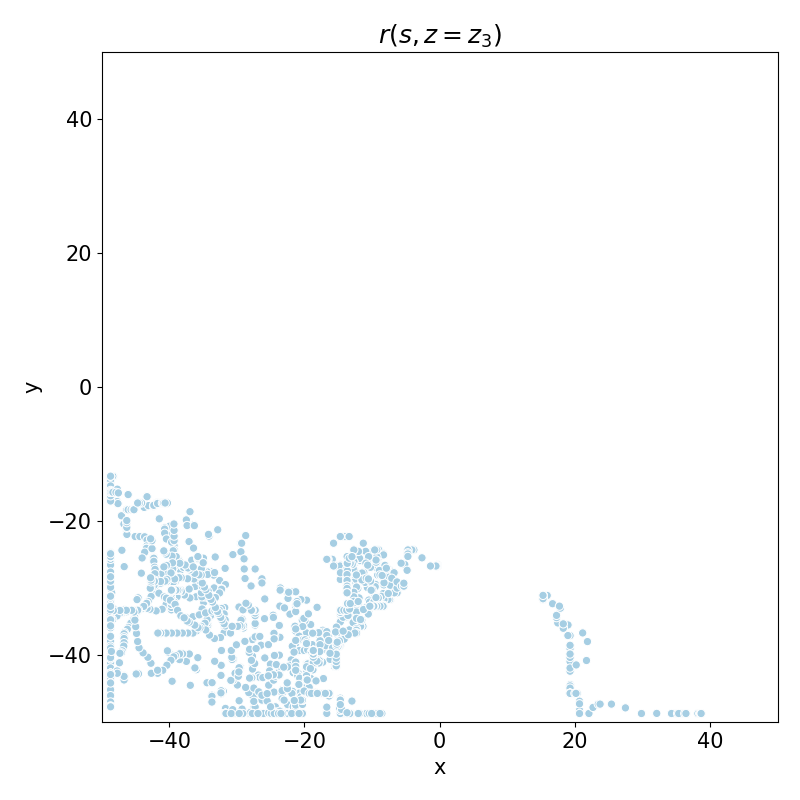}    
        \caption{}
    \end{subfigure}
    \begin{subfigure}[!hbt]{0.28\textwidth}
        \includegraphics[width=\textwidth]{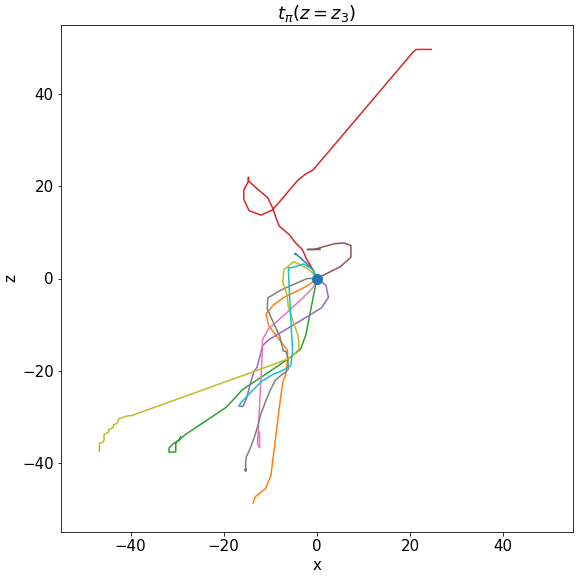}    
        \caption{}
    \end{subfigure}
    \begin{subfigure}[!hbt]{0.28\textwidth}
        \includegraphics[width=\textwidth]{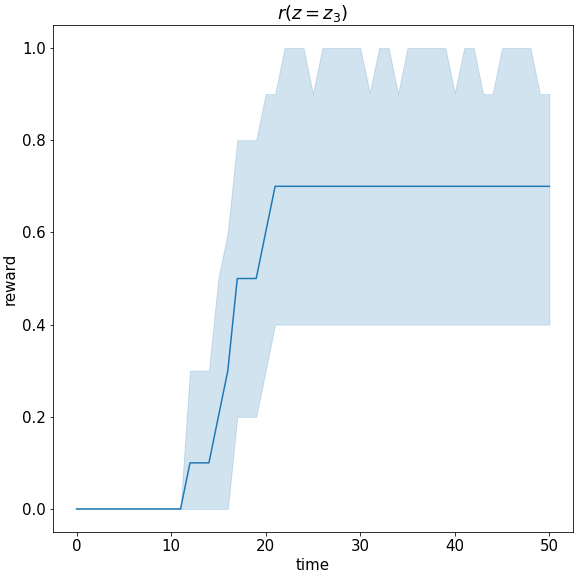}    
        \caption{}
    \end{subfigure}
    \caption{a) Observations that are encoded as the latent vector 3.
            b) and c) show the trajectories performed by the agent and the reward received in average when conditioned on skill 3. }
    \label{fig:skill3_realistic}
\end{figure*}


\subsection{Discovery and learning of skills in realistic maps}
\label{sec:realistic_maps}

The approach presented in the simple and handcrafted map of Section~\ref{ssec:map-handcrafted} fails when tested in more realistic Minecraft maps, like the one in Figure~\ref{fig:indexmaps}a.
In this case, the discovered skills depicted in Figure~\ref{fig:indexmaps}b are no longer distributed in separable clusters over the map, and one skill (in light blue) tends to dominate over the others.
Since we are treating each latent variable as a navigation goal, the ones that are spread out over the whole map will not guide but confuse the agent as it is receiving high reward in multiple places.

We overcome this problem by 
exploiting the spatial information provided by the environment in the form of coordinates. 
The learned embedding space
contains a relationship between visual features and relative spatial coordinates with respect to the initial state, always the same one. 
This allows our agent to distinguish between two visually identical mountains located at two different positions in the map.
Figure~\ref{fig:indexmaps} shows complementary skills discovered from pixels (Figure~\ref{fig:indexmaps}b) or coordinates (Figure~\ref{fig:indexmaps}c).
We would like to discover skills that take into account not only the visual similarity but also the position relative to the initial state, so we adopt a solution that considers the two types of state representations (Figure~\ref{fig:indexmaps}d).

We scale the VQ-VAE \cite{oord2017vqvae} to encode both pixels and coordinates independently. We sum up their embeddings and quantize the result to one of the codebook embeddings $e$. Now, the decoder of the model contains two independent branches, where one reconstructs an image and the other reconstructs a coordinate from the same embedding. The only difference from the original loss function is that we add a term for learning the coordinate reconstruction.

Once we have better latent embeddings for realistic maps, we can train an agent conditioned on them. Qualitative results are presented in Figure~\ref{fig:skill3_realistic}, where we can see an example of a latent encoding the bottom-left observations and the trajectories and rewards received in 10 evaluation episodes. The performance of all the conditioned policies is included in the Appendix~\ref{sec:full_realisticmap}. We observed that at least 6 skills out of 10 could be potentially leveraged in a hierarchical policy.

\begin{figure}[!hbt]
    \centering
    \includegraphics[width=0.5\columnwidth]{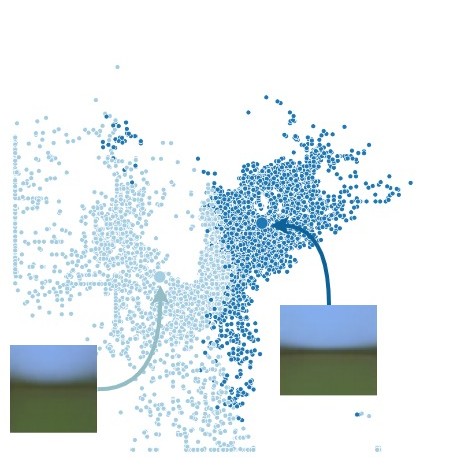}
    \caption{A very homogeneous and big region of the space becomes two different skills when using a combination of pixels and coordinates.}
    \label{fig:two_centroides}
\end{figure}

Figure~\ref{fig:two_centroides} shows a clear example of the pixels and coordinates mixing. If we based our encoder in just pixel observations we would have an almost identical latent encoding for both images. Instead, thanks to the coordinates, they become two separate latents. This may help the agent in being more precise and distinguish different regions of the space despite its visual similarity.

\section{Conclusions}
In this work we present a framework to extend the EDL paradigm to pixels. This requirement presents one main problem. Following the EDL derivation of the mutual information, we  optimize the agents based on the MSE, which does not have any notion of space in the environment. To overcome that, we propose to derive the reverse form of the mutual information between inputs and latents. In this manner we compute the intrinsic reward in the embedding space which encodes the dynamics learned during the skill-discovery phase. We observed that contrastive and variational approaches are capable of learning the mapping from inputs to latents. In the former case, we leverage the use of positive pairs based on delayed observations, and the K-Means clustering algorithm for determining the categorical latents. In the latter, we make use of a VQ-VAE model that directly allows to infer categorical latents from the input data. We qualitatively show that some skills are correctly learned, specially in the toy maps where the environment dynamics are much easier. We demonstrate that for learning useful navigation goals in realistic maps we need more information, and therefore, we learn an embedding space combining pixel and coordinates observations.


\section*{Acknowledgments}
The authors would like to thank Victor Campos and Oscar Mañas for their advise and feedback. This work was partially supported by the Postgraduate on Artificial Intelligence with Deep Learning of UPC School, and the Spanish Ministry of Economy and Competitivity under contract TEC2016-75976-R. We gratefully acknowledge the support of NVIDIA Corporation with the donation of GPUs used in this work.
\bibliography{bibliography}

\clearpage
\newpage

\onecolumn
\appendix

\section{Hyperparameters}
\label{sec:hyperparams}

\begin{center}

\begin{tabular}{ |p{4cm}|p{2cm}|  }
\hline
\multicolumn{2}{|c|}{\textbf{VQ-VAE}} \\
\hline
learning rate & 1e-3 \\
batch size & 256 \\
num hiddens & 64 \\
num residual hiddens & 32 \\
num residual layers & 2 \\
embedding dim & 256 \\
num embeddings & 10 \\
$\beta$ & 0.25 \\
decay & 0.99 \\

\hline
\end{tabular}

\begin{tabular}{ |p{4cm}|p{2cm}|  }
\hline
\multicolumn{2}{|c|}{\textbf{Contrastive}} \\ 
\hline
learning rate & 1e-3 \\
batch size & 128 \\
$\tau$ & 5e-3 \\
soft update & 2 \\
embedding dim & 128 \\
$k_\mu$ & 15 \\
$k_\sigma$ & 5 \\

\hline
\end{tabular}

\begin{tabular}{ |p{4cm}|p{2cm}|  }
\hline
\multicolumn{2}{|c|}{\textbf{Rainbow}} \\ 
\hline
learning rate & 2.5e-4 \\
batch size & 64 \\
sampling & uniform \\
max episode steps & 500 \\
update interval & 4 \\
frame skip & 10 \\
gamma & 0.99 \\
clip delta & yes \\
num step return & 1 \\
adam eps & 1.e-8 \\
gray scale & no \\
frame stack & no \\
final expl frames & 3.5e5 \\
final epsilon & 0.01 \\
eval epsilon & 0.001 \\
replay capacity & 10.e6 \\
replay start size & 1.e4 \\
prioritized & yes \\
target upd interval & 2.e4 \\

\hline
\end{tabular}
\end{center}

\clearpage
\newpage
\section{Learning from Pixels with Contrastive approach}
\label{sec:full_toyexample}
(Handcrafted map)

In this first plot, we show which observations would give a positive reward for each conditioning latent. In some cases they encode a very specific region, however, in some others the reward is spread over the map leading to poor policies.

\begin{figure}[!hbt]
    \centering
    \includegraphics[width=\textwidth]{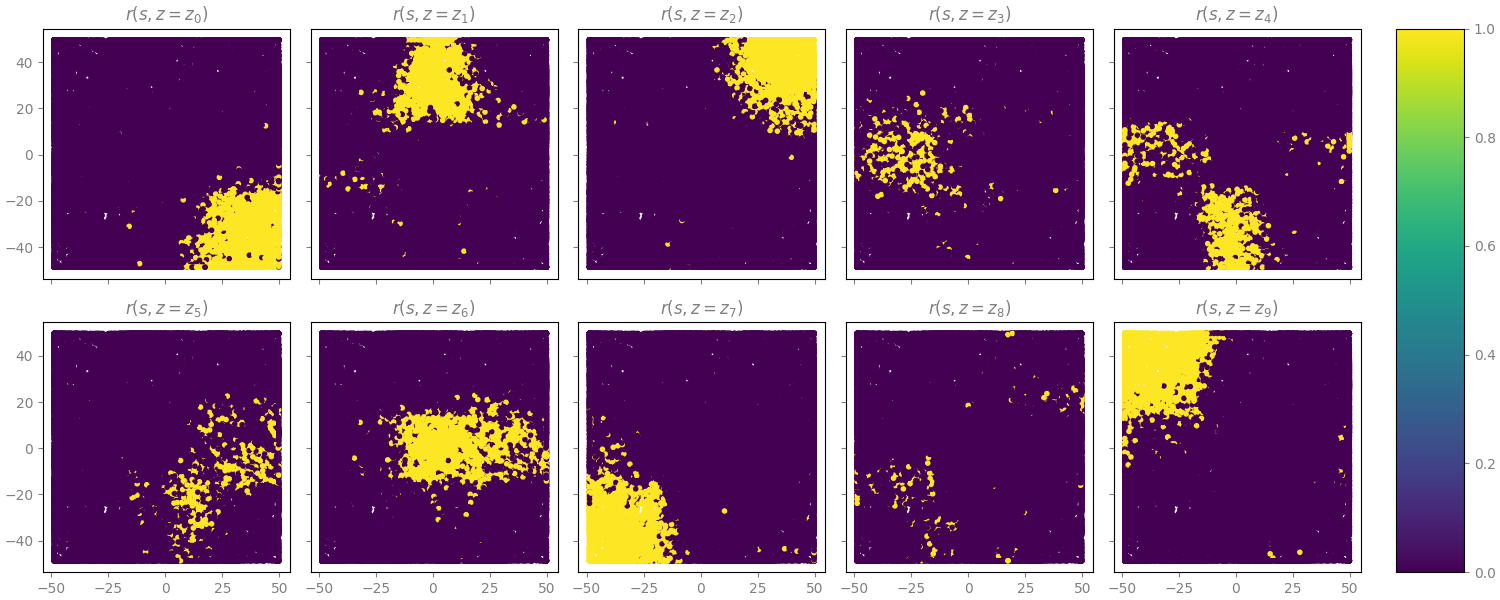}
    \label{fig:appB2}
\end{figure}

As we can see in the evaluated trajectories, latents 0,1,2,4,5,6 are properly learned. Instead, we can see that latents 3,7,8,9 are more ambiguous and the agent does not solve the self-supervised task correctly.

\begin{figure}[!hbt]
    \centering
    \includegraphics[width=\textwidth]{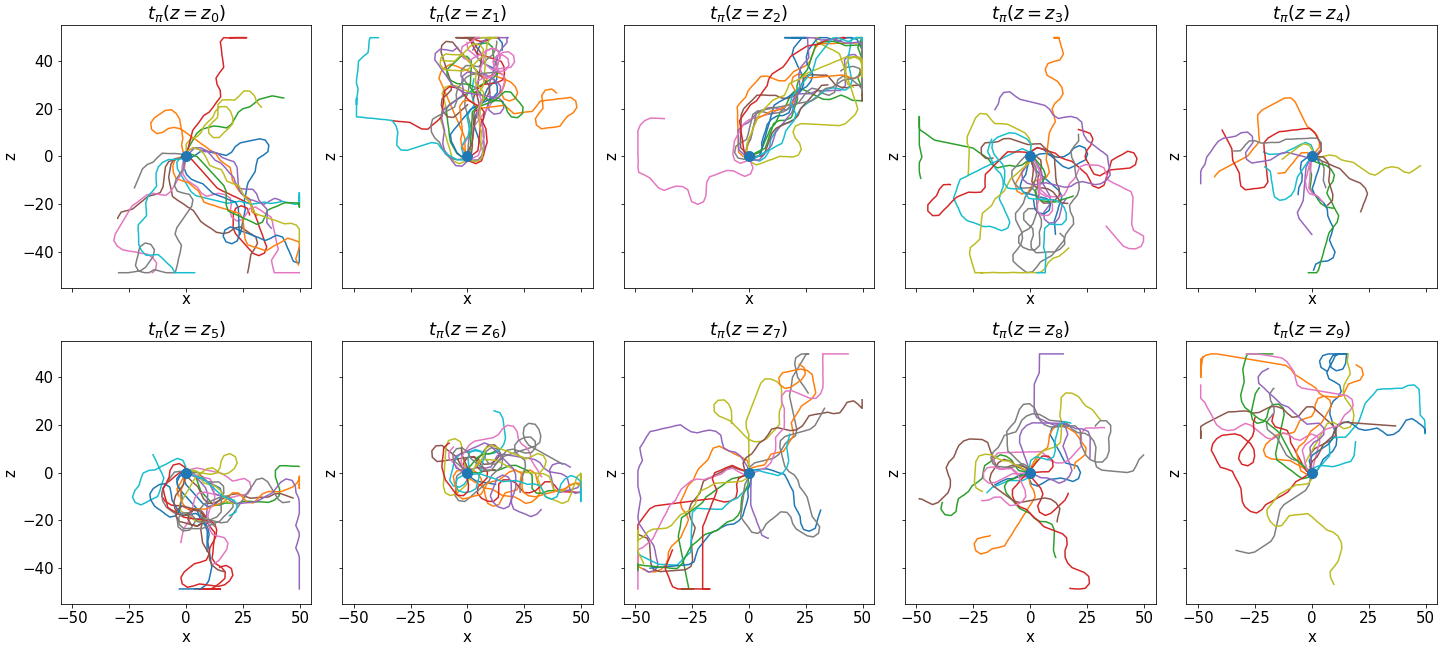}
    \label{fig:appB1}
\end{figure}

In the last plot we show the average reward of all the trajectories for each of the conditioning latents. We can see that the curves with higher reward match those trajectories in the previous plot that lead to the region encoded by the latent.

\begin{figure}[!hbt]
    \centering
    \includegraphics[width=\textwidth]{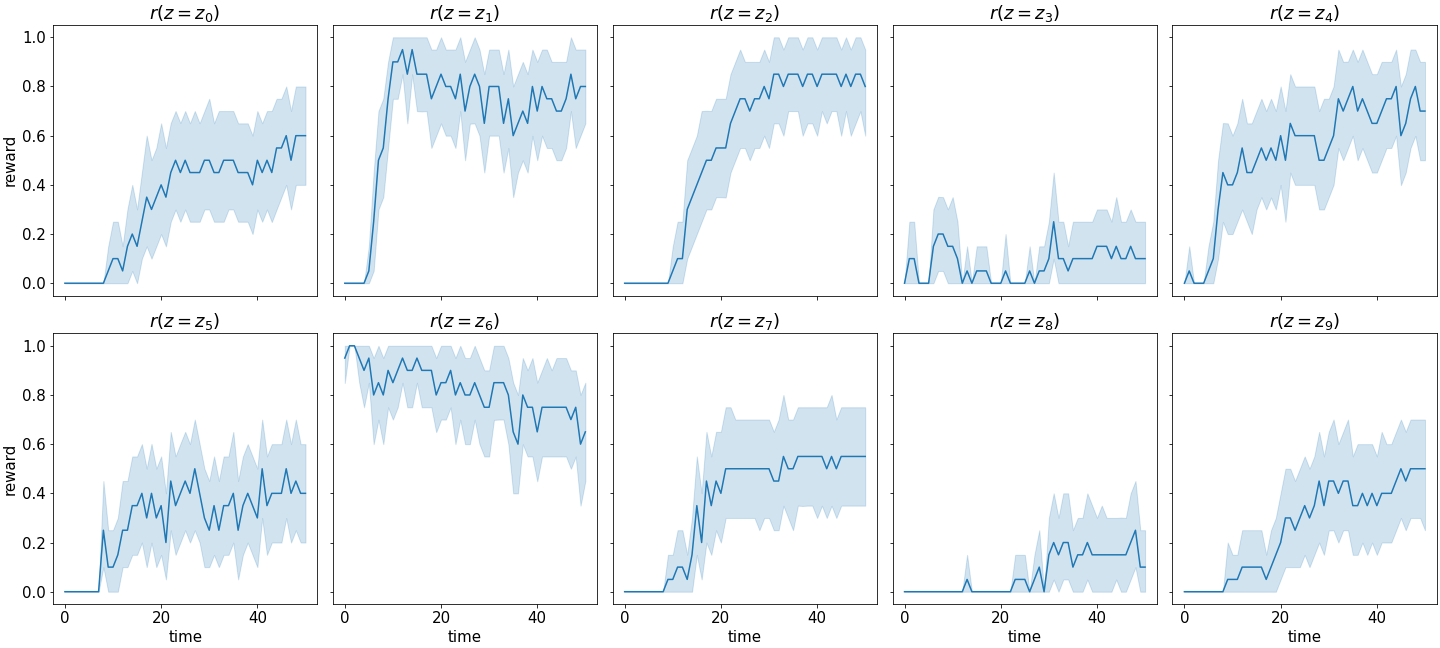}
    \label{fig:appB0}
\end{figure}

\clearpage
\newpage
\section{Learning from Pixels and Coordinates with Variational approach}
\label{sec:full_realisticmap}
(Realistic map)

In this case, we can see that despite deploying the agent in a realistic map, the discovered latents encode very specific regions of the map. This is achieved by leveraging the raw pixel information and the relative position of the agent respect to the initial point.

\begin{figure}[!hbt]
    \centering
    \includegraphics[width=\textwidth]{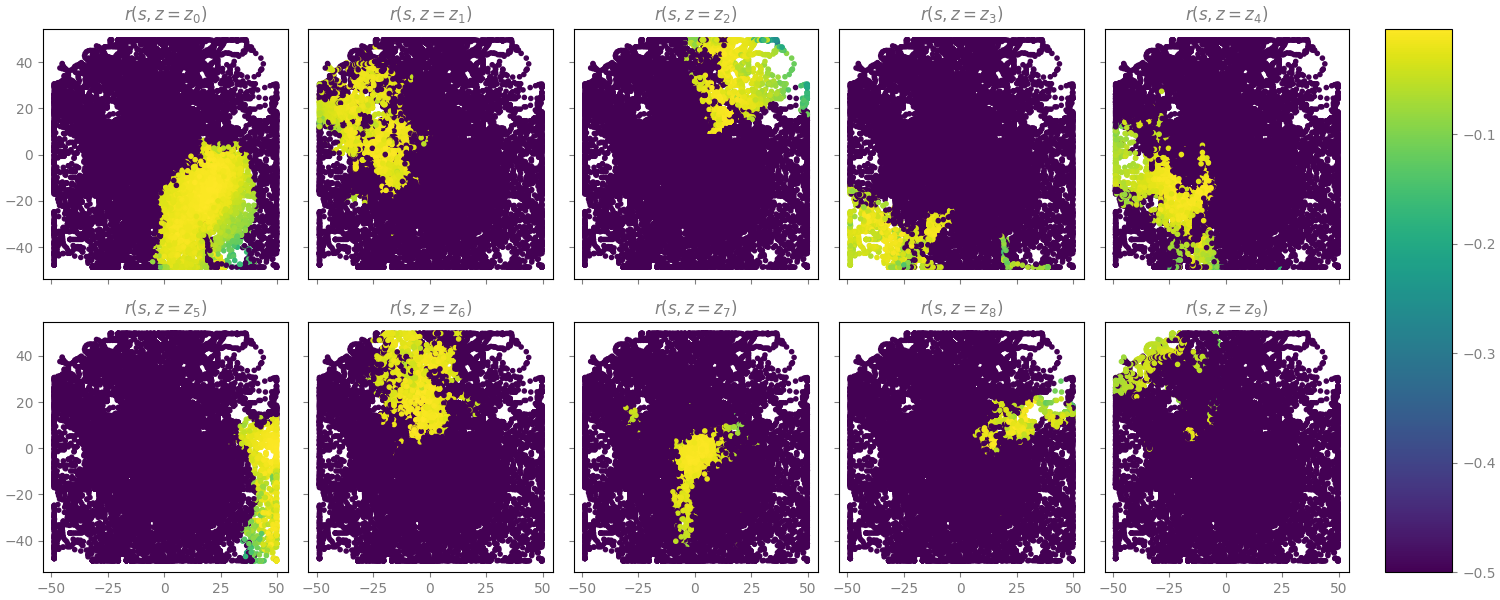}
    \label{fig:appC2}
\end{figure}

Due to the dynamics of this map, we can see that is much more difficult to learn the right policies to guide the agent towards the regions encoded by the latents. Skills 1, 2, 3, 4, 6 and 7 still achieve good results, where, regardless of the agent's initial direction, it manages to reach the desired region.

\begin{figure}[!hbt]
    \centering
    \includegraphics[width=\textwidth]{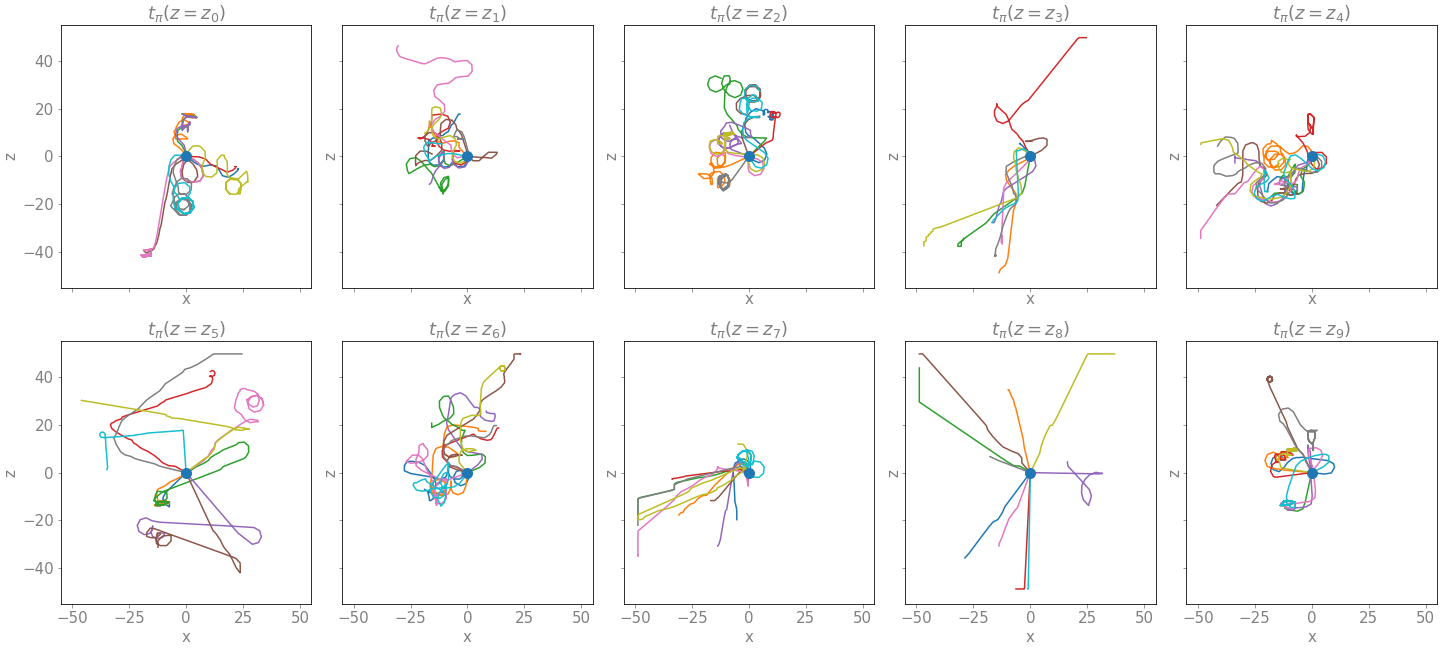}
    \label{fig:appC1}
\end{figure}

\begin{figure}[!hbt]
    \centering
    \includegraphics[width=\textwidth]{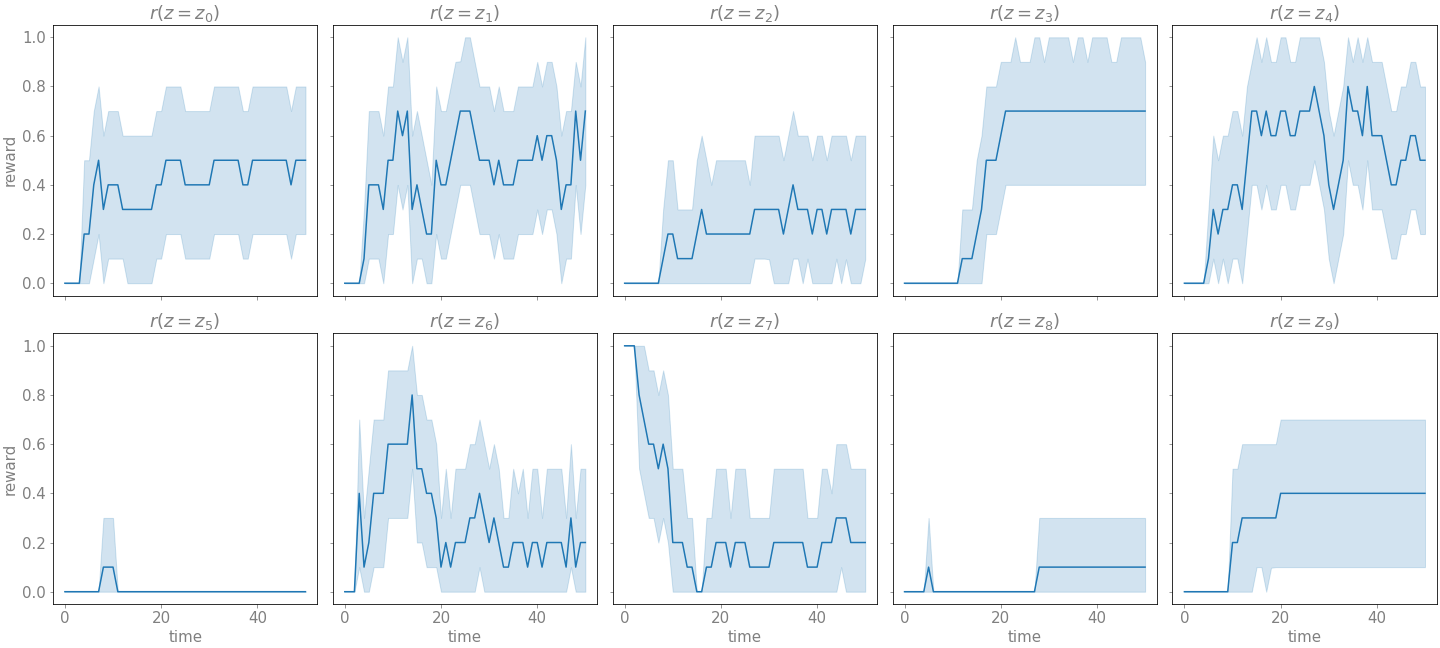}
    \label{fig:appC0}
\end{figure}

\end{document}